\documentclass[conference]{IEEEtran}
\usepackage[ruled]{algorithm2e}
\usepackage{algorithmic}

\def\ps@headings{%
\def\@oddhead{\mbox{}\scriptsize\rightmark \hfil \thepage}%
\def\@evenhead{\scriptsize\thepage \hfil \leftmark\mbox{}}%
\def\@oddfoot{}%
\def\@evenfoot{}}
\makeatother
\usepackage{amsmath}
\usepackage{amssymb}
\IEEEoverridecommandlockouts
\usepackage[space]{cite}
\usepackage{graphicx}
\usepackage{multirow,makecell}
\usepackage{paralist}
\usepackage{wrapfig}
\usepackage{amsmath}
\usepackage{multirow}
\usepackage{lscape}
\usepackage{amssymb}
\usepackage{pifont}
\newcommand\BibTeX{{\rmfamily B\kern-.05em \textsc{i\kern-.025em b}\kern-.08em
		T\kern-.1667em\lower.7ex\hbox{E}\kern-.125emX}}

\usepackage{tikz}
\usepackage{tikz-qtree}
\usetikzlibrary{trees,arrows,shapes,positioning,shadows}
\usetikzlibrary{positioning,shadows,trees,backgrounds,fit}
\pgfdeclarelayer{background}
\pgfsetlayers{background,main}
\usepackage{verbatim}
\usetikzlibrary{matrix,calc}
\usepackage{glossaries}
\include{acronyms}
 \usepackage{pgf}

\begin{document}

\title{Data-Efficient Energy-Aware Participant Selection 
for UAV-Enabled 
Federated Learning 
}

\IEEEoverridecommandlockouts

\author{
        Youssra Cheriguene, Wael~Jaafar,  Chaker Abdelaziz Kerrache, Halim~Yanikomeroglu, Fatima Zohra Bousbaa,  \\
         and Nasreddine Lagraa

\thanks{Y. Cheriguene, is with Université Amar Telidji de Laghouat, Laghouat, Algeria, and a visiting scholar at Carleton University, Ottawa ON, Canada  (e-mail: y.cheriguene@lagh-univ.dz).
{W. Jaafar is with the École de Technologie Supérieure, Montreal, QC, Canada (e-mail: wael.jaafar@etsmtl.ca).}
{H. Yanikomeroglu is with Carleton University, Ottawa, ON, Canada (e-mail: halim@sce.carleton.ca).}
C.A. Kerrache, F.Z. Bousbaa and N. Lagraa are with Université Amar Telidji de Laghouat, Laghouat, Algeria (e-mails: \{ch.kerrache, f.bousbaa, n.lagraa\}@lagh-univ.dz).
}}
\vspace{-1cm}


\maketitle

\begin{abstract}
Unmanned
aerial vehicle (UAV)-enabled edge federated learning (FL)
has sparked a rise in research interest as a result of the massive and heterogeneous data collected
by UAVs, as well as the privacy concerns
related to UAV data transmissions to edge servers.
However, due to the redundancy of UAV collected data, e.g., imaging data, and non-rigorous FL participant selection, the convergence time of the FL learning process and bias of the FL model may increase.
Consequently, we investigate in this paper the problem of selecting UAV participants for edge FL, aiming to improve the FL model's accuracy, under UAV constraints of energy consumption, communication quality, and local datasets' heterogeneity. We propose a novel UAV participant selection scheme, called data-efficient energy-aware participant
selection strategy (DEEPS), which consists of selecting the best FL participant in each sub-region based on the structural similarity index measure (SSIM) average score of its local dataset and its power consumption profile. 
Through experiments, we demonstrate that the proposed selection scheme is superior to the benchmark random selection method, in terms of model accuracy, training time, and UAV energy consumption. 
\end{abstract}

\section{Introduction}
In 5G and beyond communication networks, edge computing is considered as one of the enabling technologies for low-latency and mission-critical applications \cite{Mseddi2019,Dieye2022}. In contrast to edge servers at fixed locations, flying unmanned aerial vehicles (UAVs)-enabled mobile edge computing (MEC) plays a vital role in providing wide-area flexible deployment \cite{Jaafar2021_IoT,Cherif2020_Glob,Cherif2021_ICC}, reliable communication \cite{Jaafar2020_TVT,Jaafar2020_OJVT,Cherif2021}, data collection \cite{Safwan2019_Glob,Ghdiri2021}, and mobile MEC services \cite{2,3}.
Due to their imagery capabilities, UAVs' collected imaging data can be exploited for a plethora of applications, such as
traffic monitoring \cite{Masmoudi2021,8}, road extraction \cite{6}, and remote sensing \cite{5}. These applications need fast and real-time data analysis, thus requiring to build customized and secure machine learning (ML) models. 


Traditional ML models, which are based on the aggregation of the UAVs' sensing data at a central entity, e.g., an edge server, may raise serious privacy and data misuse concerns, due to the broadcast nature of wireless communications and the vulnerability of the central server. Moreover, transmitting raw data streams, such as UAV images/videos, to the edge server, consumes a lot of the wireless bandwidth and may saturate it at peak times. Also, such behavior may rapidly drain the UAV's battery \cite{7}. 

To tackle this issue, federated learning has been recently proposed as a promising distributed ML paradigm \cite{9}. FL allows a group of participants to collaboratively train their ML models without disclosing their local data within their communications with the central server, whereas the latter aggregates the received non-critical data to train a global model and send updates to the FL participants. Typically, not all available devices participate in the FL process. Indeed, only a subset is selected each time to participate in FL \cite{9}. This approach allows handling heterogeneous devices with different ML capabilities and data properties. Although random participant selection is a common strategy, it leads often to weak FL performances.

In addition, in the context of UAV systems, redundancy in collected data occurs often and in different ways, such as exact and near-duplicate images acquired by different UAVs.
Redundancy has a negative impact as it causes an unnecessary usage of storage space, computation resources, and power. 
Moreover, it may lead to biased Ml models
due to the lack of new and diversified information in the samples. Hence, this leads to a degradation in the model's performance. 
For instance, authors of \cite{DBLP:journals/corr/abs-1902-00423} noticed a high redundancy in datasets CIFAR-10, CIFAR-100
and ImageNet. Through experiments, they proved that the classification accuracy drops between 9\% and 14\% compared to the original redundancy-free model. Hence, it is critical to carefully select FL participants with high-quality and very low-redundant datasets.

 
Aiming to improve the performance of the UAV-enabled FL model,
we propose here a novel participant selection scheme, a.k.a., data-efficient energy-aware participant selection strategy (DEEPS),
that prioritizes participants with high data diversity and sufficient battery capacity to handle local training. To the best of our knowledge, this work is among the firsts to investigate strategic FL UAV participant selection, with respect to data and UAV constraints. Our main contributions can be summarized as follows:  

\begin{enumerate}
    \item We set up the UAV-enabled edge FL system model, including the FL model, the UAV communication, mobility, and energy consumption models, and the UAV edge computing model.
    \item Following the sub-division of the UAVs' operating area into sub-regions, we propose a novel FL UAV participant selection scheme, based on the structural similarity index measure (SSIM) average score of its local dataset and its power consumption profile.
    

    \item Through experiments on the Fire Luminosity Airborne-based Machine learning Evaluation (FLAME) dataset \cite{qad6-r683-20}, we prove the superiority of our selection scheme, in terms of FL model accuracy, training time, and UAV energy consumption, compared to the random selection benchmark. 
    
\end{enumerate}    

This paper is organized as follows.
Section II examines the related work. The system model is introduced in Section III, while the proposed FL participant selection scheme is provided in Section IV.
Then, performance evaluation is conducted in Section V. Finally,
Section VI concludes the paper. 

\section{Related Works}
\label{sec: relatedwork}
FL is an emerging paradigm that only recently it started to be investigated in the context of UAV networks.
In \cite{105}, a serverless architecture for UAV networks, a.k.a., decentralized FL for UAV networks (DFL-UN) was proposed. 
It consists of sharing
the UAV local model with the one-hop UAV neighbors.

Decentralized serverless approaches converge relatively fast when the number of participants is small, i.e., their scalability is very limited, as discussed in \cite{105}. To tackle this issue and ensure distributed FL scalability, participant selection protocols are essential.
For instance, authors of \cite{5} proposed a blockchain-based collaborative UAV-based FL architecture to securely exchange local model updates. Then, UAVs' privacy is preserved by applying
local differential privacy.
Through simulations, the proposed approach is proven to effectively improve utilities for UAVs, ensure high-quality model sharing, and guarantee FL privacy protection.
However, this approach did not take into account the UAVs' power constraints in the participants' selection process.  
In \cite{116}, battery-constrained federated edge learning (FEEL) in UAV-enabled Internet of Things (IoT) is investigated, in which UAVs can modify their running CPU frequencies to extend battery life and avoid falling out of FL training prematurely. This system is further optimized to reduce a linear combination of latency and energy consumption, by jointly allocating the computational resources and wireless bandwidth based on a deep deterministic policy gradient (DDPG) strategy.
Simulation results show that the proposed method outperforms the benchmark ones, in terms of FL convergence, latency, energy consumption, and system cost.
Authors in \cite{DBLP:journals/corr/abs-2201-01230} proposed a semi-supervised FL (SSFL) framework for privacy-preserving UAV image
recognition, where a federated
mixing (FedMix)
strategy is used to improve the naive combination of FL and semi-supervised learning under the scenarios of labels-at-client and labels-at-server. To alleviate the statistical heterogeneity problem, the federated frequency (FedFreq) aggregation rule is proposed, which adjusts
the weight of the corresponding local model according to the client's participation frequency in training.
In \cite{NASSER2022108672}, a heterogeneous wireless communication architecture is proposed for collaborative private medical analytics where UAVs utilize the information gathered by individual end-users (EUs) via wearables and environmental sensors.
In FL, an asynchronous weight update mechanism is proposed to prevent repeated learning and conserve the UAVs' and EUs' limited networking and computing resources.

\section{System Model}
\begin{figure}[t]
\centering
\includegraphics[trim={2cm 3cm 3cm 3cm},clip,width=0.999\linewidth]{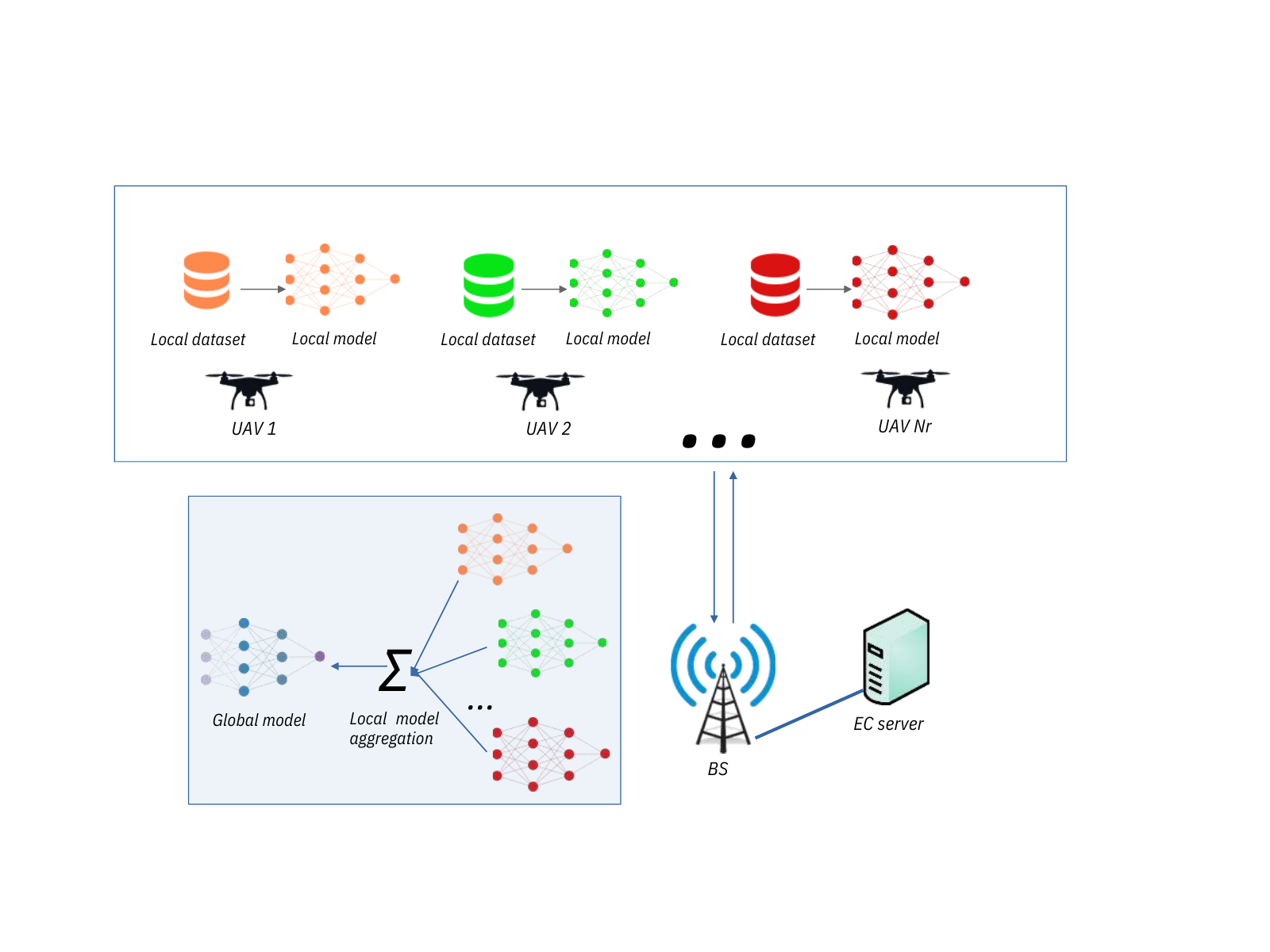}
\caption{System model.
}
\label{fig:syst}
\end{figure}

Our system model is depicted in Fig. \ref{fig:syst}.  
It is composed of several regions, where in each region $r$ a BS is paired to an edge computing (EC) server to serve ground users and also $N$ UAVs deployed in the BS's geographic area. The UAVs can also act as edge nodes, but with lower capabilities than the EC server. For the sake of simplicity, we assume that UAVs are uniformly distributed on the horizontal plane while hovering at the same altitude. Finally, EC management is realized by a regional edge controller hosted within the BS.





A user in a given region can request a service relative to any region $r$. The service is characterized by its type that describes the FL task, maximum number of global FL rounds \textit{$n_{r}^{\rm max}$}, and required number of UAVs \textit{$N_{r}$} to participate in the FL task.
Indeed, we assume that the data required to execute the user's request is collected by the UAVs in their region of interest $r$, thanks to their on-board equipment, including high-resolution cameras, GPS, LiDAR, etc., and that the latter are participants in FL training. Then, the EC server associated with the corresponding BS acts as the FL aggregator.





\subsection{FL Model}
FL is a distributed ML technique used to train a shared model while maintaining the participants' privacy.
Instead of sending raw data to an EC server, FL participants, in our case UAVs, can train their datasets locally.
The EC server gathers the local models from participating UAVs to generate a global model. 
For a learning task, each one of the participating $N_{r}$ UAVs collects a set of data to create the training dataset $d_{r}^{u}$ with size $|d_{r}^{u}|$, where $u$ identifies UAV $u$ and $|\cdot|$ is the cardinality operator. UAV $u$ trains its own neural network (NN) model with a subset of $d_{r}^{u}$, denoted $d_{r,k}^{u}$, during $T_e$ epochs (equivalent to FL round $k$, with $k=1,\ldots,n_r^{\max}$), then sends back its model update $\omega_u$ to its EC server to train the global model $\omega$.
The global model is updated by aggregating the incoming updates. Then, it transmits the generated model parameters to the participating UAVs for
another training round. These steps are repeated until the model converges or $n^{\max}_{r}$ is reached. Finally, the result is sent back to the user that triggered the request. If the latter lies within a different region than the UAVs, the result is sent between the regions through the wired BSs' backbone, e.g., fiber optic cables.



Typically, the loss function captures the model's error on training samples.
Let \textit{$F_{u}$} be the local loss function for the participating UAV $u$ and \textit{$f_s$} is the loss function on a single data sample $s \in d_{r,k}^u$. Consequently, in round $k$
\begin{equation}
\small
\label{2}
F_{u}(\omega_{u})=
\frac{1}{|d_{r,k}^{u}|} \sum_{s\in d_{r,k}^{u}} f_{s}(\omega_{u}).
\end{equation}
For the global model, the loss function in round $k$ can be expressed by
\begin{equation}
\small
\label{eq:2}
        F(\omega)=\sum_{u=1}^{N_{r}} \frac{|d_{r,k}^{u}|}{|d_{r,k}|} F_u(\omega_u),
\end{equation}
where $d_{r,k}=\cup_{u=1}^{N_{r}}d_{r,k}^{u}$. The objective of the local and global models is the minimize the loss function by optimizing and updating the NN parameters.






\subsection{Communication Model}

We assume that the UAVs are hovering in their respective regions in a non-turbulent environment, e.g., no wind, when a task is requested. 
With collected $d_{r}^{u}$ data, training for $T_e$ epochs is run by UAVs. Afterward, NN update parameters are sent to the corresponding BS/EC server through the wireless channel. For the sake of simplicity, we assume that wireless channels are reciprocal, and follow the air-to-ground free-space path-loss model, i.e., the channel coefficient between UAV $u$ and BS $b$ is written as
\begin{equation}
  \small  \label{eq:channel}
    h_{ub}=h_{bu}=\sqrt{\beta_0} d_{ub}^{-\frac{\alpha_{ub}}{2}},\; \forall u, \forall b,
\end{equation}
where $\beta_0$ is the reference channel gain, $d_{ub}$ is the 3D distance between UAV $u$ and BS $b$, and $\alpha_{ub}$ is the path-loss component expressed as
\begin{equation}
 \small   \label{eq:pathloss}
    \alpha_{ub}=\frac{a_1}{1+a_4 \exp\left({a_3 \left( \theta_{ub} - a_4 \right) }\right)}+a_2, \; \forall u, \forall b,
\end{equation}
with $a_1,\ldots,a_4$ constants defined in \cite{Shafique2021} and $\theta_{ub}$ is the elevation angle between the BS and UAV. Given that $P_{b,u}$, $P_u$, and $W_u$ are the power of BS $b$ to transmit to UAV $u$, UAV $u$'s transmit power, and reserved bandwidth to communicate to/from UAV $u$, respectively, then the channels' capacities for FL data exchange can be given by
\begin{equation}
\small
    R_{u,b}= W_u \log_2 \left(1 + \frac{P_u |h_{ub}|^2}{W_u \sigma^2} \right),\; \forall u, \forall b,
\end{equation}
and 
\begin{equation}
\small
    R_{b,u}= W_u \log_2 \left(1 + \frac{P_{b,u} |h_{bu}|^2}{W_u \sigma^2} \right),\; \forall u, \forall b,
\end{equation}
where $\sigma^2$ is the unitary power of the additive white Gaussian noise (AWGN).




\subsection{FL Latency}
FL training requires running a number of global training rounds, for which local training and data transfer between participating UAVs and an associated EC server occur. 
Each UAV spends time to train its local model for $T_e$ epochs, thus, training time, denoted $t_{r}^{u}$, can be written as
\begin{equation}
\small
\label{eq:comp}
    t_{r}^{u}=T_e \frac{\kappa  |\bar{d}_{r}^{u}|}{\gamma_u}, \; \forall u, 
\end{equation}
where $|\bar{d}_{r}^{u}|=\frac{|d_r^u|}{n_r^{\max}}$, assuming that the sizes of the training subsets are equal for UAV $u$. $\kappa$ is the number of CPU-cycles needed to process the samples, 
and $\gamma_u$ is the CPU-frequency of the UAV. 
Assuming negligible transmit time over the fiber links, negligible aggregation time at the EC server, and orthogonal access to the wireless communication channel, then the communication times from/to UAV $u$ and associated BS/EC server are given by
\begin{equation}
\small
\label{eq:tx1}
    t_{u,b}=\frac{M z}{R_{u,b}},
\end{equation}
and 
\begin{equation}
\small
\label{eq:tx2}
    t_{b,u}=\frac{M z}{R_{b,u}},
\end{equation}
where the number and size of local NN parameters are denoted by $M$ and $z$, respectively.
Consequently, the duration of one global FL round at the EC server associated with BS $b$ is
\begin{equation}
 \small   t_{r}^{b}=\max_{u \in \mathcal{U}_r} \left(t_{r}^{u}+t_{u,b}+t_{b,u}\right),
\end{equation}
where $\mathcal{U}_r$ is the set of participating UAVs in FL training.






\subsection{FL Energy Consumption}


Typically, a UAV spends most of its energy flying/hovering, while a smaller amount is dedicated to other services such as communication and computing. For the sake of simplicity, we ignore the energy consumption related to flying and focus only on the energy consumed for FL\footnote{This assumption is acceptable since we assumed that UAVs are hovering in a non-turbulent environment, thus the hovering power is a constant for the same type of UAVs.}. 
The UAV's consumed energy for one global FL round is composed of the energy needed to train the local model, denoted $E_{r}^{u}$, in addition to the energy for transmissions $E_{u,b}$, where
\begin{equation}
\small
    E_{r}^{u}= \chi_u t_{r}^{u} {\gamma_{u}}^3,  
\end{equation}
and
\begin{equation}
\small
    E_{u,b}= P_u t_{u,b},
\end{equation}
with $\chi_u$ is the energy consumption of the UAV's CPU chips \cite{116}.
 Since the BS/EC has an unlimited source power, we ignore its consumption for communication and computing.






\section{Problem Formulation}
On one hand, correlated data may potentially reduce the accuracy of a model.
In certain instances, the model may overfit the correlated data, which has a negative impact on the accuracy of each UAV participant $u$.
On the other hand, the energy usage of each UAV is limited by its battery capacity, which would reduce the number of participations in FL.
Our objective is to trade-off between two conflicting objectives, namely improving the FL accuracy, while reducing energy consumption.
Let $\textbf{x}=[x_{1,k},\ldots,x_{N,k}]$ be the vector of binary variables that indicate the participation or not of UAVs from the set of a region $r$ in a training round $k$, where $N$ is the number of available UAVs. Also, let $B_{u,k}$ be the residual battery capacity of UAV $u$ at FL round $k$, and $B_{\textrm{max}}$ is the maximal battery capacity of any UAV. 
The problem of selecting the best participants while accounting
for energy consumption and local accuracy, denoted $A_u$, can be formulated as follows: 

\begin{subequations}
\small
	\begin{align}
	\label{p1}
	\max_{\mathbf{x}} & \sum_{u=1}^{N} x_{u,k} \left(\xi A_u+(\xi-1) (B_{u,k}/B_{\textrm{max}})\right), \tag{$\mathcal{P}1$} \\
	\label{cc1}
	\text{s.t.}\quad 
      &  x_{u,k} (E_{r}^{u}+E_{u,b}) \leq B_{u,k}, \; \forall u=1,\ldots,N, \tag{$\mathcal{P}1$.a}\\ 
  &  x_{u,k} \in \{0,1\},  \forall u-1,\ldots,N, 
  \tag{$\mathcal{P}1$.b}\\
  &  \sum_{u=1}^N x_{u,k} = N_r, 
  \tag{$\mathcal{P}1$.c}
	\end{align}
\end{subequations}  
where $\xi$ and $(\xi-1)$ are wighting factors of the sub-objectives, with $\xi \in [0,1]$. ($\mathcal{P}1$.a) refers to the energy consumption constraint, while ($\mathcal{P}1$.b) explains the binary nature for participants selection. Finally, ($\mathcal{P}1$.c) ensures that $N_r$ UAVs participate in each FL round.  
Problem $\mathcal{P}1$ is complex to solve due to the intractable expression of the accuracy. Thus, we opt in what follows for a heuristic approach, where a modification of the objective is needed.

\section{Proposed Participant Selection Strategy: DEEPS}
\label{solution}



According to \cite{zhu2021federated}, UAVs with high-quality datasets, i.e., with less correlated data and adequate training sample size, can result in efficient local model updates and faster FL convergence.  
Based on that, we propose a novel participant selection strategy, called DEEPS, aiming to optimally select UAVs for FL training. 

To provide high-quality local datasets, we propose to select UAVs in distinct sub-regions of $r$, while taking into account their battery capacity to sustain FL training up to convergence or to $n_r^{\max}$ FL rounds. For the sake of simplicity, we assume that the space of region $r$ is partitioned into 3D sub-regions, denoted $\{sr_1,\ldots,sr_M\}$  ($M$ is the number of sub-regions), that host different groups of UAVs.
Moreover, we assume an FL task that relies on datasets built from the UAVs' onboard camera-captured images.
However, the accuracy $A_u$ in ($\mathcal{P}1$) depends on the quality of used datasets in training. 
UAVs may collect data frequently, thus generating  large amounts of data with redundant and correlated information. 

To attenuate the impact of this phenomenon on the accuracy performance, we propose to use the structural similarity index measure (SSIM) average score to measure the similarity or dissimilarity between images in a given dataset. Conventionally, the Euclidean distance is the basic tool used to evaluate the correlation of two images $\Gamma_1$ and $\Gamma_2$. 
It can also be expanded to determine the similarity of images with affine pixel intensities \cite{6046605}.
However, if images $\Gamma_1$ and $\Gamma_2$ are for instance two views of the same scene, i.e., obtained from different viewing angles, the assumption of an affine relationship between image intensity values no longer holds, thus resulting in a more complex relationship between the pixel intensities. Moreover, the choice of the right image similarity metric requires a compromise between image processing speed and efficiency.

In such a context, SSIM is advocated as the most accurate similarity metric \cite{1284395}. Indeed, SSIM measures the visual quality and by extension, the perceptual proximity between images based on subjective quality assessments through the analysis of vast databases. The SSIM is more accurate than other metrics, such as the mean squared error (MSE) and the peak signal-to-noise ratio (PSNR), since it is based on the comparison of structural information from images rather than pixel-wise error used by MSE and PSNR \cite{1284395}.    


SSIM is made up of 
three components, namely the visual impact of changes in image brightness, contrast, and any remaining defects, together known as structural alterations.
When combined together with the SSIM function, SSIM can be calculated as 
\begin{equation}
\small
\label{13}
\text{SSIM}(\Gamma_1,\Gamma_2)=
 \frac{(2 \mu_{\Gamma_1}\mu_{\Gamma_2}+c_{1})(2 \sigma_{\Gamma_1\Gamma_2}+c_2)}{(\mu_{\Gamma_1}^2 + \mu_{\Gamma_2}^2+c_1) (\sigma_{\Gamma_1}^2+ \sigma_{\Gamma_2}^2+c_2) }, 
\end{equation}
where $\mu_{\Gamma_1}$ and $\mu_{\Gamma_2}$ represent the mean intensities of images $\Gamma_1$ and $\Gamma_2$, respectively. The standard deviations (square roots of variances) are $\sigma_{\Gamma_1}$ and $\sigma_{\Gamma_2}$, and they reflect estimates of the signals' contrast. Finally, $\sigma_{\Gamma_1\Gamma_2}$ is the co-variance of the two images, whereas $c_1$ and $c_2$ are two stabilizers acting on a weak denominator \cite{1284395}.
The values range of SSIM is 0 to 1, where 1 means that images $\Gamma_1$ and $\Gamma_2$ are identical, while 0 means that they are completely different.

Therefore, in order to improve the effectiveness of FL learning, we propose to consider SSIM of the training subsets as a metric to maximize $A_u$. Hence, we formulate a novel problem $\mathcal{P}2$, where  

\begin{subequations}
\small
	\begin{align}
	\label{p2}
	\max_{\mathbf{x}} & \sum_{u=1}^{N} x_{u,k}\left(\xi \left(1-\text{SSIM}\left(d_{r,k}^{u}\right)\right)+(\xi-1)(B_{u,k}/B_{\textrm{max}})\right), \tag{$\mathcal{P}2$} \\
	\label{cc_1}
	\text{s.t.}\quad 
& sr_i \cap \{ \text{UAV }  u \;|\; x_{u,k}=1 \} \neq \emptyset, \; \forall i=1,\ldots,M \tag{$\mathcal{P}2$.a}  \\
  & \text{($\mathcal{P}1$.a)}-\text{($\mathcal{P}1$.c)}, \tag{$\mathcal{P}2$.b) -- ($\mathcal{P}2$.d}
	\end{align}
\end{subequations}
where $(1-\text{SSIM}(d_{r,k}^u))$ refers to the diversity (or dissimilarity) of data within subset $d_{r,k}^u$, and ($\mathcal{P}2$.a) guarantees that at least one UAV participant is selected from each sub-region.
Problem $\mathcal{P}2$ is a mixed-integer nonlinear programming (MINLP) problem, which is NP-hard. To solve it, we propose a heuristic method, as detailed below.

Specifically, we propose the following UAV participant selection method. First, the BS/EC server initiates selection by broadcasting a request to UAVs in all sub-regions to calculate their datasets' similarity scores with SSIM.

Subsequently, in order to solve Problem \ref{p2}, the EC server ranks the UAVs of each sub-region in a descending order based on the utility function that linearly combines SSIM measure, in order to reflect local dataset diversity, and the remaining battery level, i.e., UAV $u$ score in FL round $k$ is
\begin{equation}
\small
\label{eq:crit}
    S_{u,k}=\xi (1-\text{SSIM}(d_{r,k}^u)+ (1-\xi) \frac{B_{u,k}-E_r^u-E_{u,b}}{B_{\textrm{max}}}, \;\forall u.
\end{equation}
Next, UAVs with the highest $S_{u,k}$ in each sub-region are selected to participate in the FL round $k$. 
Selected UAVs preprocess their subset. Specifically, through the comparison of the subset samples one by one, a sample is discarded if it presents high SSIM above a threshold SSIM$_{th}$, when compared to another one. 
By doing so, quasi-redundant data is eliminated and the storage capacity of the UAV is better managed.




Given the received training parameters and preprocessed data, FL training can be run until convergence or reaching $n_r^{\max}$. It is to be noted that at the beginning of each FL round, participating UAVs are selected/re-selected in order to avoid involving UAVs with battery shortages in the FL process.
Furthermore, the proposed solution may be used for FL tasks with different restrictions.
For instance, in applications where energy saving is a priority, a smaller $\xi$ is preferred, whereas an increased $\xi$ value could recommended in FL tasks where accuracy is more critical than energy savings. 
The proposed DEEPS selection method is summarized in Algorithm \ref{Alg3}.

\begin{algorithm}[t]{
  \algsetup{linenosize=\small}
  \small
  	
		\SetAlgoLined
		\textbf{Input:} number of UAVs $N$, number of participants $N_r$, number of global rounds $n_{r}^{\rm max}$, 
  $\xi$ utility weight parameter \; 
  
		\textbf{Output :} $\omega$ global model after $n_{r}^{\rm max}$ rounds of training \; 
				
		\underline{\textbf{Steps}}\\
      EC server starts the initial global model \;

	
 

\For{round $k=1,\ldots, n^{r}_{\rm max}$}{
\For{each SubRegion \text{in} Region $r$}{ 
   \For{each \text{UAV} in SubRegion}{ 	

    Calculate SSIM($d_{r,k}^u$) using eq. (\ref{13}) \;
    Update $S_{u,k}$ using eq. (\ref{eq:crit}) \; 

    Order UAVs in \textit{SubRegion} from highest to lowest $S_{u,k}$ 
    }
Select $N_r$ UAVs from the \textit{SubRegions} with the highest $S_{u,k}$ scores \; 
 \For{each UAV $u\in N_r$}{
		 \If{ Subset $d_{u,k}^{r}$ is not preprocessed}{
		Remove redundant data samples from $d_{u,k}^{r}$ using the SSIM$_{th}$ threshold\;
  }
  	Train local model         $\omega_u$ \;
        Upload $\omega_u$ and $B_u$ to the EC server \;
		}
 
 }
 Aggregate received local weights and update $\omega$ \;
 }
 }
	\caption{DEEPS}
	\label{Alg3}
\end{algorithm}

\section{Experiments and Results}
\label{sec: experiment}

\subsection{Dataset and Preprocessing}
\begin{figure}[t]
\includegraphics[width=0.99\linewidth
]{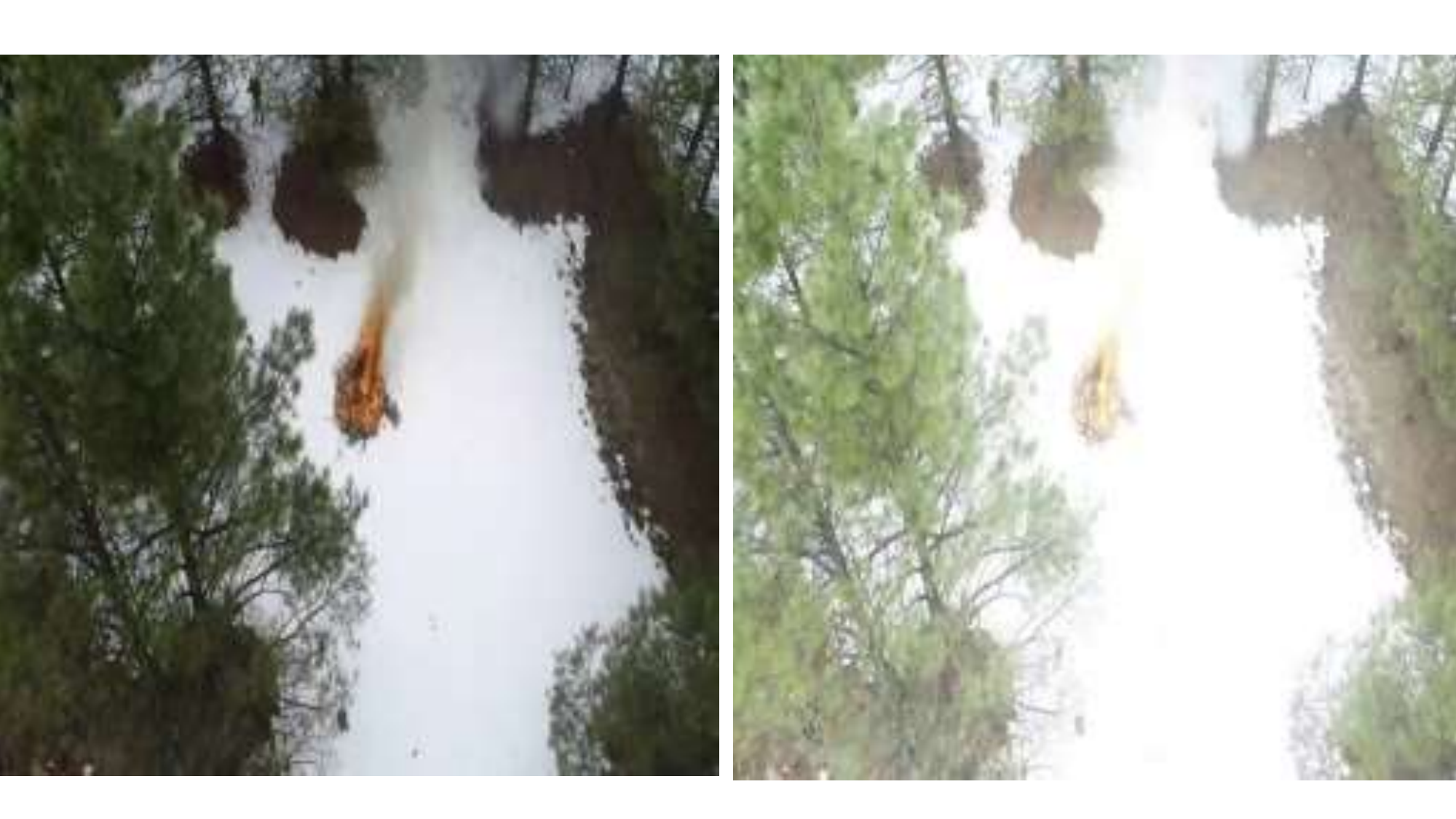}
\caption{Comparison of two images from FLAME dataset.}
\label{appR2}
\end{figure}

The FL experiments conducted here are related to an application that requires captured images from UAVs, such as in a disaster situation where a scene needs to be assessed to identify damage, survivors, etc. 
Specifically, we relied on the public FLAME dataset \cite{qad6-r683-20}.
The latter is composed of images taken by UAVs
during a controlled pile burn in Northern Arizona, USA. It contains images with fire (fire images) and others without any fire (non-fire images). Also, it includes continuous frames taken from videos captured by the UAVs, where the images demonstrate significant similarities. An example is illustrated in Fig. \ref{appR2} for which SSIM$=0.85$.  

To reflect the sub-region partitioning, UAVs that are in the same sub-region have
similar viewpoints from the FLAME dataset and each local dataset is split into 80\% for training and 20\% for testing.
Moreover, federated learning performance is influenced by many sorts of data distributions from clients. 
To exhibit the non-i.i.d. behavior expected of FL datasets, we vary the size of local datasets between 500 and 1000 for each UAV with a uniform distribution of samples in each class.

In this experiment, SSIM ensures that contiguous frames in a local dataset have enough dissimilarities.
Hence, we have defined three different methods for selection, namely random selection (equivalent to randomly picking UAV participants), DEEPS based on SSIM$_{th}=0.5$, and DEEPS based on SSIM$_{th}=0.1$. 



\subsection{Experimental Setup}
We assume a region $r$ consisting of a BS/EC server serving $10$ aerial sub-regions. For the sake of simplicity, we assume that $N$ UAVs are uniformly distributed in these sub-regions with batteries $B_u \in [10^3,10^4]$ Joules (J), $\forall u=1,\ldots, N$. 
Similarly to \cite{116}, we assume that $\gamma_u=10$ MHz, $\kappa=7 \times \; 10^4$ CPU cycles, $\chi_u=10^{-22}$ Watt, $P_u=0.28$ Watt, and $P_{b,u}=1$ Watt, $\forall u=1,\ldots,N$. We consider that diversity in the local datasets of participants is equally important to the residual energy for the FL task $\tau_r$, therefore,  we set $\xi=0.5$.  Based on DEEPS, we define a first scenario, called \textit{Scenario 1}, where $N_r=10$ UAVs will be selected from $N=40$ UAVs in each round to participate in FL training for at most $n_r^{\max}=200$ rounds. We assume here that a single UAV is selected from each sub-region.
A second scenario (\textit{Scenario 2}) is designed where $N=100$ and $N_r=20$ UAVs are selected among them, where 2 UAVs are selected from each sub-region.






The aggregator and UAV FL participants use the same convolutional neural network (CNN) model. Each participant trains its model to perform an FL classification task (presence or absence of fire) in the images. Specifically, 
the binary classification model used here is a small version of Xception network \cite{8099678} \cite{Chris2017}, i.e., a deep CNN architecture that involves depthwise separable convolutions. 
The CNN model is composed of an input layer, two hidden layers, and an output layer. The input layer is of size $250\times250\times3$, which reflects the size of input images in pixels and the three red-green-blue (RGB) color channels. The RGB values are scaled to float values between 0 and 1.
Hidden layers are composed of 2-dimensional (2D) convolutional blocks with a size of 8 and a stride of $2\times 2$, where each block is followed by a Rectified Linear Unit (ReLU) and batch normalization \cite{9209059}. 
ReLU is utilized as the activation function to increase the network's fitting ability, while batch normalization equalizes the inputs to each layer for each mini-batch to accelerate the learning process. 
The output layer is a fully-connected one with the Sigmoid function to
normalize the output to a probability distribution. The learning rate is set to $10^{-2}$. Finally, the neural network is trained by the Adam
optimizer, and the cross-entropy loss
function is adopted to measure the classification performance. Finally, the batch size is set to 32.

\begin{figure}[t]
    \includegraphics[trim={1.1cm 6.5cm 2.1cm 7.5cm},clip,width=0.97\linewidth]{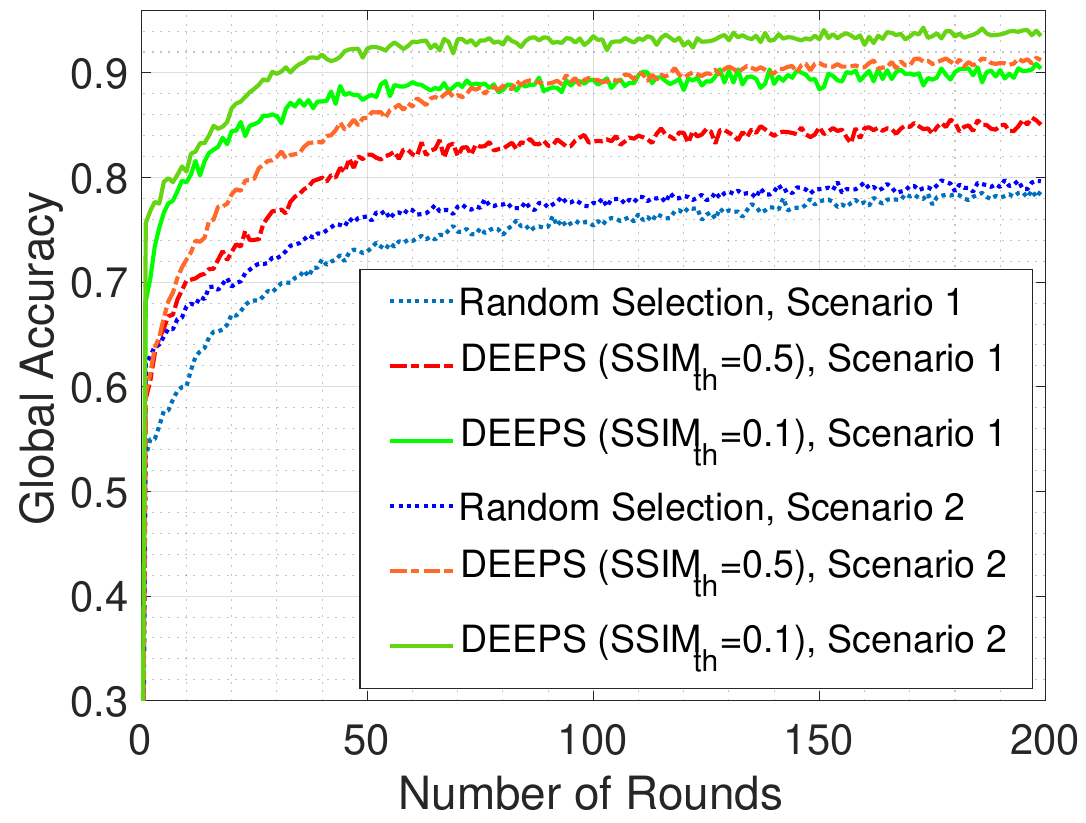}
\caption{Global accuracy vs. number of FL rounds.}
\label{acc}
\end{figure}

Fig. \ref{acc} compares the performance of the proposed selection strategy DEEPS to those of the random selection benchmark, in terms of global accuracy, for both scenarios 1 and 2. Random selection exhibits the worst accuracy performance, mainly due to the high risk of training with correlated samples in both scenarios, especially when participants are selected from the same sub-region or from UAVs with highly redundant data. 
In contrast, DEEPS is superior due to its efficient preprocessing based on SSIM and sub-region-based UAV participant selection. Also, a lower SSIM$_{th}$ improves further the training accuracy that reaches 90\% for SSIM$_{th}=0.1$, against 85\% for SSIM$_{th}=0.5$ and 77\% for random selection in Scenario 1. 

In Scenario 2, higher performances are achieved with the preference for DEEPS with SSIM$_{th}=0.1$. Indeed, the availability of a higher number of UAVs and selecting two UAVs per sub-region favors a more accurate FL training without delaying convergence. 
Finally, we notice that generally, DEEPS achieves convergence faster than random selection in both scenarios.

\begin{table}[t]
\caption{Performance comparison for different participant selection methods}
    \centering

    \begin{tabular}{|c*{4}{c}|}

\hline \textbf{Scenario} & \textbf{Selection Criteria} &  \makecell{\textbf{$\lambda_t$}\\ \textbf{(s)}} & \makecell{\textbf{$\chi_r$}} & \makecell{\textbf{$\rho_t$} \\\textbf{(min)}} \\ 
\hline

       \multirow{1}{*}{Scenario 1} & Random Selection& 120.33 &  123 & 245.9\\ 
          &DEEPS (SSIM=0.5) & 58.21 & 97 & 93.71\\
          
          &\textbf{DEEPS (SSIM=0.1)} & \textbf{51.02}& \textbf{88} & \textbf{77.46}\\ 
          \hline

       \multirow{1}{*}{Scenario 2} & Random Selection & 137.89& 153 & 343.4\\
          
          &DEEPS (SSIM=0.5) & 70.55 & 115 & 131.2\\
          
          &\textbf{DEEPS (SSIM=0.1)} & \textbf{63.85}& \textbf{96} & \textbf{78.61}\\
          \hline
          
    \end{tabular}

\label{tab:2}
\end{table}

Table \ref{tab:2} illustrates the performance of DEEPS against the benchmark selection method in terms of average round time $\lambda_t$, number of rounds until convergence $\chi_r$, and elapsed time until convergence $\rho_t$.
In the first scenario, DEEPS executes each FL round in less than 60 seconds, for any SSIM$_{th}$ value, which is half of the required time by random selection to complete a single FL round.
Indeed, this is due to the preprocessing step in DEEPS that reduces the dataset size and guarantees the use of only informative samples. In contrast, random selection chooses UAVs with unprocessed datasets and potentially with redundant information. A similar conclusion can be drawn for Scenario 2, where the convergence time is slightly higher than in Scenario 1. This is explained by the increased number of UAV participants in the FL system. 

\begin{figure}[t!]
    \includegraphics[trim={0cm 5cm 1.2cm 5cm},clip,width=0.97\linewidth]{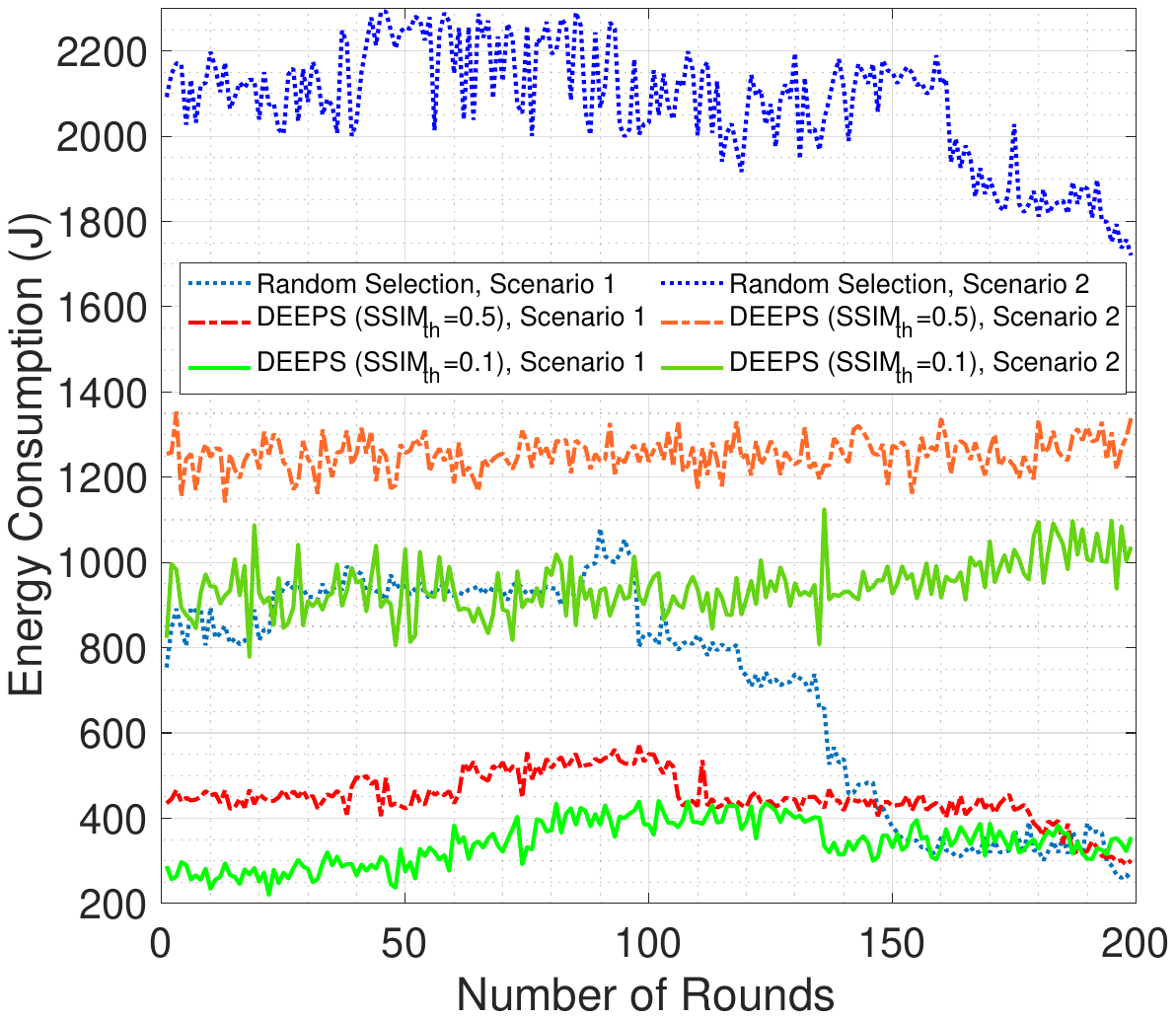}
    \label{energy1}
\caption{Energy consumption vs. number of FL rounds.}
\label{energy}
\end{figure}


In Fig. \ref{energy}, we present the energy consumed by participating UAVs of DEEPS and random selection benchmark, in joules (J).
For Scenario 1, the energy consumption is high at first, then it drops after a number of FL rounds for any participant selection method, e.g., for random selection, energy consumption drops by 900 J after 100 rounds. This is caused by the high battery drainage that forces a number of UAVs to drop from FL training. 
Nevertheless, UAVs using DEEPS consume the smallest amount of energy, e.g., about 400 J (resp. 300 J) for SSIM$_{th}=0.5$ (resp. SSIM$_{th}=0.1$), with a slight drop after the 105$^{th}$ (resp. 132$^{th}$) round by 76 J (resp. 30 J) for SSIM$_{th}=0.5$ (resp. SSIM$_{th}=0.1$). Indeed, data preprocessing and selection of UAVs with higher $B_{u,k}$ levels in DEEPS ensure that FL training consumes less energy and UAVs last longer in FL than random selection. In contrast, random selection has to execute training with more data, in addition to using UAVs that might have limited battery capacities.
Similar conclusions can be drawn for Scenario 2 where we notice a stable energy consumption for DEEPS compared to the benchmark. However, energy consumption is higher than for Scenario 1 due to the involvement of a higher number of UAVs in FL training.

\vspace{-8pt}
\section{Conclusion}
\label{sec: Conclusion}
In this paper, we investigated the problem of UAV participant selection for FL, while taking into account the datasets' similarity/dissimilarity and energy consumption. 
By adopting SSIM similarity score, applying dataset preprocessing, sub-dividing the UAVs' 3D region, and defining a UAV score function, we designed a novel participant selection method, called DEEPS. Through experiments, we demonstrated the superior performances of DEEPS compared to the random selection benchmark, in terms of FL accuracy, training time, and energy consumption. In future work, we will focus on the optimal partitioning of the UAVs' 3D region and on FL operation with UAV mobility and heterogeneous equipment.


\vspace{-5pt}
\bibliographystyle{IEEEtran}
\bibliography{bibfile}
\end{document}